%% file: main.tex
\documentclass[conf]{new-aiaa}

\usepackage{graphicx}
\usepackage{amsmath}

\usepackage{amssymb}
\usepackage{algorithm}
\usepackage{algpseudocode}
\usepackage{booktabs}
\usepackage{array}
\usepackage{siunitx}
\usepackage{xcolor}
\hypersetup{colorlinks=true,allcolors=blue}

\usepackage{tikz}
\usetikzlibrary{arrows.meta,positioning,calc,fit,backgrounds,shapes.geometric,shadows.blur}

\definecolor{ink}{HTML}{1F2933}
\definecolor{cAgent}{HTML}{0E8A82}
\definecolor{cAgentBg}{HTML}{E0F2F0}
\definecolor{cStage}{HTML}{2C6FBB}
\definecolor{cStageBg}{HTML}{EAF1FA}
\definecolor{cCred}{HTML}{D98A1E}
\definecolor{cCredBg}{HTML}{FBF0DC}
\definecolor{cStop}{HTML}{B5483F}
\definecolor{cStopBg}{HTML}{F7E6E3}
\definecolor{cWin}{HTML}{2E8B57}
\definecolor{cWinBg}{HTML}{E6F2EB}
\definecolor{cOrch}{HTML}{6C5B7B}
\definecolor{orbit}{HTML}{AAB6C0}

\newcommand{\code}[1]{\texttt{\small #1}}

\newcommand{\cwhLLM}{\SI{0.101}{\meter}}            
\newcommand{\cwhLLMsd}{\SI{0.007}{\meter}}          
\newcommand{\cwhRand}{\SI{0.635}{\meter}}           
\newcommand{\cwhFloor}{\SI{4.04}{\meter}}           
\newcommand{\cwhFloorSd}{\SI{0.26}{\meter}}         
\newcommand{\cwhSigmaMargin}{15.0}                  
\newcommand{\cwhDelta}{\SI{3.94}{\meter}}           
\newcommand{\cwhDeltaCI}{[3.75,\,4.12]\,\si{\meter}}
\newcommand{\cwhIters}{57}                          
\newcommand{\cwhRandN}{60}                           

\newcommand{\rpoLLM}{0.38}           
\newcommand{\rpoLLMsd}{0.07}         
\newcommand{\rpoFloor}{8.45}         
\newcommand{\rpoFloorSd}{1.55}       
\newcommand{\rpoSigmaMargin}{5.2}    
\newcommand{\rpoDelta}{8.07}         
\newcommand{\rpoDeltaCI}{[6.97,\,9.18]}
\newcommand{\rpoSingle}{0.33}        
\newcommand{\rpoRand}{6.46}          
\newcommand{\rpoRandN}{24}           
\newcommand{\rpoIters}{23}           
\newcommand{\rpoTarget}{0.7}         

\title{Agentic AutoResearch for Space Autonomy: An Auditable, LLM-Driven Research Agent for Aerospace Control Problems}

\author{Amit Jain\footnote{Postdoctoral Associate, Department of Aeronautics and Astronautics.}
  \ and Richard Linares\footnote{Associate Professor, Department of Aeronautics and Astronautics.}}
\affil{Massachusetts Institute of Technology, Cambridge, MA, 02139}

\begin{document}

\maketitle

\begin{abstract}
\input{sections/00_abstract}
\end{abstract}

\input{sections/01_introduction}
\input{sections/02_related_work}
\input{sections/03_methodology}
\input{sections/04_problems}
\input{sections/05_results}
\input{sections/07_conclusion}

\bibliography{references}

\end{document}

%% file: sections/00_abstract.tex
Spacecraft guidance, navigation, and control functions are increasingly realized
as learned policies distilled from expert solvers. Developing such a policy is
itself a research process: an investigator selects an architecture and
hyperparameters, runs experiments, and must determine whether an apparent
improvement is genuine or merely seed noise. This paper presents AutoResearch, a
framework in which a large language model autonomously drives that loop for aerospace
control problems, coupled with a credibility layer, built into the loop, that
certifies each reported result against the problem's own measured seed noise. The
language model serves only as the offline research agent that develops the control
policy; the trained policy it produces is then deployed onboard the spacecraft, while
the model itself never operates the vehicle. At each iteration the agent reads a
plain-language problem description and the run history, proposes a single edit to the
training script, executes it, and logs the outcome. No reported result is credited
until it passes the same three checks: measured per-problem seed noise, reseeded
verification of the best configuration, and leave-one-out pruning of the agent's
edits. The same loop is applied, unchanged, to two aerospace control problems: a
Clohessy--Wiltshire relative rendezvous and a safety-constrained collision-avoidance
docking past a keep-out zone, each calibrated against a known optimal-control
benchmark. In both, the audited policy clears the measured seed noise by many
standard deviations; an undirected search over the same parameters does not. On the
docking problem the gap becomes categorical: undirected search yields no feasible
policy, while the learned policy stays outside the keep-out zone on every seed. The contribution is the autonomous,
auditable research process itself: a reusable means of developing learned control
policies and certifying that their reported gains are genuine.

%% file: sections/01_introduction.tex
\section{Introduction}
\label{sec:intro}

Spacecraft autonomy increasingly depends on learned components. Guidance,
navigation, and control functions that were once specified entirely by hand,
among them rendezvous and proximity-operations planners, and powered-descent
guidance laws, are now frequently realized as policies trained
from data or distilled from expert solvers~\citep{sanchez2018realtime}. Building
such a component is an iterative research process in its own right: the
investigator picks an architecture and hyperparameters, runs experiments,
diagnoses failure modes, and decides whether an apparent improvement is a real
gain or just seed-to-seed variation. The loop is slow and labor-intensive, and
when it is run informally it invites over-claiming, since a single favorable run
is easily mistaken for a robust result, a failure mode well documented across
machine learning~\citep{henderson2018deeprl,agarwal2021precipice}.

The recent emergence of large language models (LLMs) that can read source code,
reason about it, and act on it raises a concrete question for the aerospace
community: can the research loop itself be automated, and can it be automated in
a way that remains transparent, reproducible, and honest? Recent LLM-driven
research agents show that autonomous experimentation is
feasible~\citep{lu2024aiscientist,romeraparedes2024funsearch}, and LLMs have been
used directly as optimizers~\citep{yang2024opro}. But speed without statistical
discipline is what the reproducibility literature warns against: an agent that
runs experiments faster only compounds the risk of mistaking noise for progress.

This paper takes up that question for aerospace control problems and presents
\emph{AutoResearch}, an agentic framework in which an LLM autonomously proposes,
executes, and analyzes machine-learning experiments, paired with a credibility
layer that lives inside the loop rather than being bolted on afterward.
Throughout, the LLM serves strictly as the research agent that drives
experimentation: it reads run histories, forms hypotheses, and selects the next
configuration to try. It operates offline, within the research loop, and produces
the trained control policy that ultimately flies the spacecraft. The autonomy under
study is thus that of the research process itself.

\paragraph{Contributions.} This paper makes three contributions.
\begin{enumerate}
\item \textbf{A reusable framework for agentic experimentation.} The framework defines a
\emph{family} contract: a plain-language description, one editable training script
with a delimited hyperparameter region, a single structured metric, and an
append-only run log. One LLM-driven agent loop then applies unchanged across
aerospace problems whose physics differ sharply (Sec.~\ref{sec:method}).
\item \textbf{A credibility layer that makes autonomous results trustworthy.}
The credibility layer integrates measured per-problem seed noise, reseeded verification of the
agent's best configuration, and leave-one-out pruning of its individual edits,
applied uniformly so that a real improvement is separated from seed luck and the
contribution of each edit is made explicit (Sec.~\ref{sec:credibility}).
\item \textbf{Audited results on two aerospace control problems.} The same loop runs,
unchanged, on a Clohessy--Wiltshire relative rendezvous and a
safety-constrained collision-avoidance docking past a keep-out zone, each a
calibrated problem whose optimal solver fixes a best-achievable benchmark and whose
fixed baseline configuration sets the scale of measured
noise. In both, the audited policy clears that measured seed noise by many standard
deviations; a parallel undirected search over the same parameters trails it, and does
so decisively on the docking problem, where it produces no feasible policy at all
while the learned policy holds the hard safety constraint and stays clear of the
keep-out zone on every seed. Leave-one-out pruning then shows which of the agent's
edits actually carry each result. The novelty is the autonomous research loop and its
audit taken together, the process that both produces the policy and certifies it
(Secs.~\ref{sec:problems}--\ref{sec:results}).
\end{enumerate}

The remainder of the paper reviews related work (Sec.~\ref{sec:related}),
describes the framework and its credibility layer (Sec.~\ref{sec:method}), states
the two demonstration problems and reports their audited results
(Secs.~\ref{sec:problems}--\ref{sec:results}), and closes with limitations and
future work (Sec.~\ref{sec:conclusion}).

%% file: sections/02_related_work.tex
\section{Related Work}
\label{sec:related}

This work draws on several lines of research and is best understood by how it
departs from each. The first is the use of large language models (LLMs) as
autonomous agents that conduct, rather than merely assist, research. The AI
Scientist~\citep{lu2024aiscientist} closes the full loop of idea generation,
experimentation, analysis, and writing for small machine-learning studies, and
FunSearch~\citep{romeraparedes2024funsearch} pairs an LLM with an automated
evaluator and evolutionary search to discover programs that improve on known
mathematical constructions. A closely related thread treats the LLM itself as the
optimization operator: OPRO~\citep{yang2024opro} states the optimization task in
natural language and has the model propose successively better solutions from a
prompt of prior trials and their scores, while OptFormer~\citep{chen2022optformer}
learns a transformer-based universal hyperparameter optimizer from large tuning
corpora. These systems establish that an LLM can drive an end-to-end search or
discovery loop. What they do not settle is whether a reported improvement is
statistically real or merely a favorable random seed, the question the
credibility layer is built to answer.

A fast-growing literature now builds LLM \emph{agents} that autonomously run
machine-learning experiments rather than merely optimize a scalar.
MLAgentBench~\citep{huang2023mlagentbench} casts ML experimentation as an agent
task of reading and writing code, executing it, and iterating on a metric;
AIDE~\citep{jiang2025aide} frames ML engineering as a tree search over candidate
programs; AutoML-Agent~\citep{trirat2025automlagent} orchestrates a multi-agent
pipeline from data to deployment; and the MLE-bench
benchmark~\citep{chan2025mlebench} measures such agents against human Kaggle
leaderboards. Closer to the hyperparameter-tuning core of the present loop,
AgentHPO~\citep{liu2024agenthpo} has an LLM propose configurations, read the
resulting trials, and iterate, an agentic successor to OPRO. Most relevant in
spirit is Curie~\citep{kon2025curie}, which likewise seeks \emph{rigor} in
autonomous LLM experimentation, but enforces it \emph{procedurally}, through
experimental-setup validation, controlled design, and reproducible, re-runnable
workflows. The credibility layer here is complementary, and \emph{statistical} where
Curie is procedural: it denominates each reported number in a measured per-problem
seed noise and gates it on reseeded verification and leave-one-out ablation of
the agent's own edits. A recurring lesson from this cluster, made explicit by
MLE-bench, is that such agents are high-variance and ought to be scored over many
seeds; this work takes that lesson inside the loop, turning a reporting convention into an
acceptance gate.

Automated hyperparameter search, of course, predates LLMs, and it supplies the
baseline adopted here. \citet{bergstra2012random} established random
search as both an effective method and the natural baseline against which any
adaptive search must be measured, and modern frameworks such as
Optuna~\citep{akiba2019optuna} provide efficient define-by-run search and pruning.
This paper adopts that baseline directly: each LLM campaign here is
run against a random search over the same editable surface under a matched
per-iteration budget, anchored to a shared first iteration so both arms start from
a common configuration.

Seed sensitivity and under-powered evaluation are recurring hazards in machine
learning. \citet{henderson2018deeprl}
showed that deep reinforcement-learning results are highly sensitive to random
seeds and implementation details; \citet{agarwal2021precipice} argued that
evaluation in the few-seed regime must report uncertainty rather than point
estimates; and the NeurIPS reproducibility program~\citep{pineau2021reproducibility}
codified reporting practices to combat the same failure. Together these works
motivate the central design choice of this paper: an agent that experiments faster only
compounds the over-claiming risk unless its conclusions are filtered through
measured seed noise, reseeded verification, and ablation of its own edits, so that
filtering is built directly into the loop.

The problems on which the framework is exercised come from learned guidance,
navigation, and control for spacecraft. Behavioral cloning has a long lineage in
control, from ALVINN~\citep{pomerleau1988alvinn} to the no-regret analysis of
DAgger~\citep{ross2011dagger}, and has been used to distill optimal guidance
solvers into fast feedback policies, as when \citet{sanchez2018realtime} represent
optimal landing guidance with deep networks for onboard evaluation. The control
formulations relied on here are classical: relative orbital motion via the
Clohessy--Wiltshire equations~\citep{clohessy1960terminal}, convex powered-descent
guidance~\citep{acikmese2007convex}, and safety enforcement through control
barrier functions~\citep{ames2019cbf}. They also build on the authors' prior work in
optimal feedback control and stochastic reachability
analysis~\citep{jain2023hjb,jain2025hjmpc,jain2023reachability,jain2023stochastic,jain2026sparse};
this reachability analysis underlies the screening of feasible initial conditions in the present demonstration problem. On the learning side,
reinforcement learning has been applied directly to spacecraft guidance and
control, as in deep reinforcement learning for six-degree-of-freedom planetary
landing~\citep{gaudet2020landing} and adaptive guidance with reinforcement
meta-learning~\citep{gaudet2020metarl}, and the authors have previously unified multi-phase
trajectory optimization in a single transformer-based reinforcement-learning
policy~\citep{jain2025multiphase}.

Most relevant, and most important to distinguish from the present work, is a
recent line that places language models inside the spacecraft control loop. LLMs
have been used as autonomous spacecraft operators that act on natural-language
prompts~\citep{rodriguez2024lmoperators}, vision-language models have been cast as
operator agents in the space domain~\citep{carrasco2025vlm}, a reasoning LLM
trained with group relative policy optimization has been used to synthesize
stabilizing control policies~\citep{jain2026grpo}, and compact networks applied
recursively have served as efficient optimal controllers~\citep{jain2026trc}. In
each of these the model, or a network it trains, is the controller. The approach here
inverts that arrangement: the LLM never flies or commands the vehicle; it is the
offline research agent that proposes, runs, and analyzes the experiments which
produce and validate such controllers.

These threads have not previously been brought together. LLM agents now routinely
automate machine-learning experimentation~\citep{huang2023mlagentbench,jiang2025aide,chan2025mlebench},
and a recent line pursues experimental rigor~\citep{kon2025curie}; the
reproducibility literature shows that fast experimentation without statistical
discipline is dangerous~\citep{henderson2018deeprl,agarwal2021precipice}; and the
aerospace-learning literature supplies well-posed, safety-critical control problems
whose learned components are just the sort a research agent would tune. To the authors'
knowledge, though, no prior system gates each reported result through an in-loop
statistical credibility audit of measured seed noise, reseeded verification, and
leave-one-out ablation of the agent's own edits, and none has been applied to
aerospace control problems with a known optimal-control benchmark. That combination
is the paper's contribution: an LLM-driven research agent whose every headline
result is so gated, demonstrated on two aerospace control families.

%% file: sections/03_methodology.tex
\section{The AutoResearch Framework}
\label{sec:method}

AutoResearch turns the machine-learning research loop into a closed,
machine-readable state machine. The human investigator specifies a problem once,
in a fixed format called a \emph{family}; an LLM-driven agent then repeatedly
proposes a configuration, runs it, reads the result, and decides what to try
next, until a stopping condition fires. A separate \emph{credibility layer}
operates offline on the recorded trail and decides whether any apparent
improvement is real, reproducible, and understood. The two are kept distinct on
purpose: the loop is free to search quickly, even optimistically, because nothing
it reports is believed until the credibility layer has audited it.
Figure~\ref{fig:loop} shows the loop, and the remainder of this section makes
each of its pieces precise.

\input{sections/fig_loop}

\subsection{Problem formalization}
\label{sec:formalization}
AutoResearch formalizes the research loop as optimization over a configuration space. A
family fixes a finite set of $d$ editable hyperparameters, each with a
declared domain, so the space of configurations the agent may explore is the
product
\begin{equation}
\Theta \;=\; \prod_{j=1}^{d} \Theta_j,
\qquad
\Theta_j \in \bigl\{\,
[\ell_j,u_j]\cap\mathbb{Z},\;\;
[\ell_j,u_j]\subset\mathbb{R},\;\;
[\ell_j,u_j]_{\log},\;\;
\mathcal{C}_j
\,\bigr\},
\label{eq:config-space}
\end{equation}
where each axis $\Theta_j$ is an integer interval, a real interval, a
log-scaled real interval, or a finite categorical set $\mathcal{C}_j$. These four
domain types are the ones the framework can sample and patch, and they
induce the natural per-axis laws used by the random search below: an integer or
real axis is drawn uniformly, a log-scaled axis as
$\exp\!\bigl(\mathcal{U}(\log\ell_j,\log u_j)\bigr)$, and a categorical axis
uniformly over $\mathcal{C}_j$. Because the agent may assign only keys that
already exist in the editable region, $\Theta$ is fixed for the duration of a
campaign; introducing a genuinely new degree of freedom is a deliberate human
act, not a move the agent can make.

Evaluating a configuration is expensive and stochastic. Training and scoring a
single configuration $\theta$ at random seed $s$ yields a scalar metric
\begin{equation}
f:\Theta\times\mathcal{S}\to\mathbb{R},
\qquad
F(\theta)\;=\;\mathbb{E}_{s\sim\mathcal{S}}\!\bigl[f(\theta;s)\bigr]
\;\approx\;\frac{1}{|\mathcal{S}'|}\sum_{s\in\mathcal{S}'} f(\theta;s),
\label{eq:objective}
\end{equation}
where $f(\theta;s)$ is one full training run and $F(\theta)$ is the seed-averaged
quantity that actually matters. The basic tension is visible already: the agent
must search on single noisy draws of $f$, while any honest claim has to be made
about the mean $F$. An orientation
$\sigma_{\mathrm{dir}}\in\{+1,-1\}$ is fixed: it equals $+1$ when larger metric values are
better and $-1$ when smaller values are better. The oriented gap
$\sigma_{\mathrm{dir}}\bigl(f(a)-f(b)\bigr)$ is then positive exactly when
configuration $a$ is better than $b$. This single sign lets every later threshold
be written once, independent of whether the family maximizes or minimizes.

The agent is then a proposal policy that maps the recorded history to the next
configuration to try,
\begin{equation}
\theta_t \;=\; \pi\!\bigl(\mathcal{H}_{t-1}\bigr),
\qquad
\mathcal{H}_{t-1} \;=\; \bigl\{(\theta_i,\, f(\theta_i; s_i),\, a_i)\bigr\}_{i=1}^{t-1},
\label{eq:policy}
\end{equation}
where $a_i$ is the structured annotation attached to proposal $i$: the
hypothesis, the axes touched, and the self-declared kill criterion described in
Sec.~\ref{sec:proposer}. When $\pi$ is a language model that reads the trail in context and emits the next
trial, the LLM acts as an optimizer~\citep{yang2024opro}. This contrasts with
amortized approaches, which learn a universal tuner offline from large corpora of
past studies~\citep{chen2022optformer}. The natural non-adaptive reference is random
search~\citep{bergstra2012random}, which ignores the history entirely and draws
\begin{equation}
\theta^{\mathrm{rs}}_t \;\overset{\text{iid}}{\sim}\; P \;=\; \bigotimes_{j=1}^{d} P_j,
\qquad
\theta^{\mathrm{rs},\star}_N \;=\;
\operatorname*{arg\,opt}_{t\le N}\, f\!\bigl(\theta^{\mathrm{rs}}_t; s_t\bigr),
\label{eq:randsearch}
\end{equation}
with $P_j$ the per-axis law of Eq.~\eqref{eq:config-space} and the optimum
oriented by $\sigma_{\mathrm{dir}}$ (a best-of-$N$ selection). The contrast
between the history-dependent policy of Eq.~\eqref{eq:policy} and the memoryless
sampling of Eq.~\eqref{eq:randsearch} is the adaptivity this paper sets out to
measure. Sequential model-based and Bayesian optimization sit between the
two, fitting a surrogate to the observed pairs and choosing each query by an
acquisition rule~\citep{shahriari2016taking,akiba2019optuna}. The LLM policy can
be read as a surrogate-free, language-conditioned member of that family. Its
practical advantage is that it reasons directly over the same source code and run
log a human researcher would.

\subsection{The family abstraction and its contract}
\label{sec:family}
A \emph{family} is a self-contained research problem, realized as a directory of
four artifacts: a plain-language description (\code{program.md}) of the goal, the
metric, the baseline, and the allowed scope of changes; a single editable
training script (\code{train.py}) implementing the dynamics, data, model, and
evaluation; a registry (\code{family.json}) holding the target metric, the
reference baseline, and the current best configuration; and an append-only log
(\code{runs.jsonl}) of every completed experiment. Because every family exposes the same contract, the
identical agent loop applies unchanged across problems with very different
dynamics. That contract rests on four invariants.

The first is a delimited editable region. \code{train.py} marks the configurable
block with an opening banner, then a series of top-level \code{NAME = value}
assignments, then a closing \code{\# Fixed Configuration} banner, and these
assignments are exactly the axes of $\Theta$ in Eq.~\eqref{eq:config-space}. The
agent may change only values inside that region, and only keys that already
exist; the dynamics, the expert solver, the dataset, and the evaluation protocol
live beneath the second banner and are immutable to it. The second invariant is a
single structured metric. On success, \code{train.py} prints exactly one line of
the form \code{FINAL\_METRIC: \{"run\_id": ..., "<metric>": <float>, ...\}}, whose
primary key matches \code{family.json}; $f(\theta;s)$ is read straight from that
line, with no free-form parsing. Auxiliary fields, such as a success rate or a
constraint-violation rate, are logged for the agent to reason about but never
change the optimization target. The third invariant keeps two
separate logs. \code{train.py} appends one row per \emph{successful}
experiment to \code{runs.jsonl}, while the orchestrator records
\emph{every} iteration, including invalid proposals, failures, and stops, to
\code{proposals.jsonl}. The success log is therefore never contaminated by dead
ends, yet the full decision trail stays reconstructable. The fourth invariant is a
read-only preparation step. A \code{prepare.py} supports a one-time, expensive
\code{--warm-cache} and a fast precondition \code{--check}; the orchestrator runs
the check before the first live experiment, and the agent never edits either.

\subsection{The agent loop}
\label{sec:loop}
The loop is built from three small, independently testable modules, summarized in
Algorithm~\ref{alg:loop}: a \emph{proposer} that wraps the LLM, a \emph{runner}
that executes a configuration, and an \emph{orchestrator} that drives the state
machine and enforces termination.

\begin{algorithm}[t]
\caption{The AutoResearch agent loop for one family.}
\label{alg:loop}
\begin{algorithmic}[1]
\Require family contract $(\Theta,\theta_0,f,\sigma_{\mathrm{dir}})$; policy $\pi$;
budgets $(t_{\max},C_{\max},R_{\max},\phi_{\max},G_{\max})$
\State $\mathcal{H}_0 \gets \varnothing$;\quad
$\mathcal{V} \gets \varnothing$ \Comment{$\mathcal{V}$: configurations already evaluated}
\State $b_0 \gets \text{undefined}$;\quad $\phi \gets 0$;\quad $t \gets 1$
\While{no trigger of $T_\star$ is active \textbf{and} $t \le t_{\max}$}
  \State $(\theta_t, a_t) \gets \pi(\mathcal{H}_{t-1})$
  \Comment{one forced tool call: propose or stop}
  \If{$a_t$ is a stop request}
     \State enforce justification gate; \textbf{break}
  \EndIf
  \If{$\theta_t \in \mathcal{V}$ \textbf{or} $\mathrm{keys}(\theta_t)\not\subseteq\Theta$}
     \State record invalid; $\phi \gets \phi+1$; \textbf{continue}
  \EndIf
  \State $y_t \gets f(\theta_t; s_t)$
  \Comment{isolated subprocess, wall-clock timeout}
  \State $v_t \gets \textsc{Verdict}(y_t, b_{t-1})$
  \Comment{Eq.~\eqref{eq:perrun}}
  \State $\mathcal{H}_t \gets \mathcal{H}_{t-1}\cup\{(\theta_t,y_t,a_t,v_t)\}$;\quad
  $\mathcal{V}\gets\mathcal{V}\cup\{\theta_t\}$
  \State update best $b_t$;\quad $\phi \gets 0$;\quad $t \gets t+1$
\EndWhile
\State \Return trail $\mathcal{H}$ \Comment{audited offline by the credibility layer}
\end{algorithmic}
\end{algorithm}

The proposer presents the model with the current state and forces it to return
exactly one of two structured tool calls, with no free-text channel to scrape.
The \code{propose\_experiment} tool requires a hypothesis, a dictionary of
hyperparameter overrides, one or more \emph{axis} labels classifying which part
of the search space the move touches, an expected mechanism, an explicit
comparison to the prior best, and a numerical failure prediction that serves as a
kill criterion written down before the run. The \code{stop\_research} tool
requires only a status, a conclusion, and the set of axes tested. Forcing structured
output makes every decision in Eq.~\eqref{eq:policy} machine-readable and removes
a common source of brittleness in LLM agents. Each turn the model is shown the same evidence a human would have, in a fixed
order: any warnings raised by the orchestrator, the full \code{program.md}, the
\code{family.json} registry, a compact summary of the random-search arm, the
\code{runs.jsonl} history, and the current values of the editable hyperparameters.
\label{sec:proposer}

To guard against a model that merely recalls a good configuration from
answer-bearing fields rather than searching, the framework supports a
\emph{blind mode}. The description is replaced by a redacted variant that states
the invariant problem facts but withholds the target and the baseline, and the
run history is reduced to (configuration, primary-metric) pairs. Blind mode is
the honest test of iterative search, and is the setting used for the campaign
reported in this paper. The framework supports both Anthropic and OpenAI
backends for $\pi$, as well as the key-free random proposer that realizes
Eq.~\eqref{eq:randsearch}. The campaigns reported in this paper use Anthropic's
Claude Sonnet~4.5.

The runner reads the editable region of \code{train.py} by matching the two
banners and extracts the top-level assignments. It applies the proposed overrides
by rewriting only the matching value expressions, preserving comments and
formatting and refusing any key absent from $\Theta$. It then executes
\code{train.py} as an isolated subprocess under a wall-clock timeout, inheriting
the allocation's GPU, and parses the single \code{FINAL\_METRIC} line from
standard output. Editing real source in place, rather than threading a
configuration object, means the agent operates on the same artifact a human
researcher would edit, and that each recorded experiment is reproducible by
re-running the patched script.

The orchestrator assembles the context, validates each proposal, invokes the
runner, updates the registry's best configuration when a run improves on it, and
writes one record per iteration to \code{proposals.jsonl}. Two of its
responsibilities are worth specifying. First, it restricts the search
to distinct points: a proposal is rejected if either its override map or the full
configuration it would produce has been seen before, so that
\begin{equation}
\theta_t \notin \mathcal{V} \;=\; \{\theta_1,\dots,\theta_{t-1}\}
\quad\text{and}\quad
o_t \notin \{o_1,\dots,o_{t-1}\},
\label{eq:dedup}
\end{equation}
where $o_t$ is the canonicalized override map of proposal $t$. The agent
therefore cannot spend budget re-evaluating a known point, and a campaign of $N$
accepted iterations explores $N$ distinct configurations. Second, it terminates
the campaign at a stopping time defined as the first iteration at which any
trigger becomes active,
\begin{equation}
T_\star \;=\; \min\Bigl\{\, t \ge 1 :\;
c_t > C_{\max} \;\vee\;
r_t \ge R_{\max} \;\vee\;
\rho_t \;\vee\;
t \ge t_{\max} \;\vee\;
\phi_t \ge \phi_{\max} \;\vee\;
g_t > G_{\max} \;\vee\;
\Pi_t
\,\Bigr\},
\label{eq:stop}
\end{equation}
where $c_t$ is elapsed wall-clock against budget $C_{\max}$, $r_t$ the number of
successful runs against a cap $R_{\max}$, $\rho_t = \bigl[\sigma_{\mathrm{dir}}
(b_t - y_{\mathrm{tgt}}) \ge 0\bigr]$ the predicate that the running best $b_t$ has
reached the family target $y_{\mathrm{tgt}}$, $t_{\max}$ the iteration cap,
$\phi_t$ the count of consecutive failed or invalid proposals against a cap
$\phi_{\max}$ (three in these runs), $g_t$ the cumulative LLM spend against a
monetary cap $G_{\max}$, and $\Pi_t$ the plateau predicate of
Eq.~\eqref{eq:plateau} below. These triggers are checked in the fixed precedence in which
they are written. The agent's own \code{stop\_research} is an additional
termination path, admitted only through a justification gate. A stop is accepted
when the target is genuinely met, or else when at least six experiments have
completed and at least four distinct axes have been exercised; a positive
close asserted without the target actually being reached is rejected outright.
Every campaign is therefore bounded in time, in number of experiments, and in
cost, and premature or over-stated closure is structurally difficult.

\subsection{The credibility layer}
\label{sec:credibility}
The components above let an agent search quickly; making its results trustworthy is
the job of the credibility layer. Deep reinforcement-learning and machine-learning
results are well known to be sensitive to random seeds and implementation
detail~\citep{henderson2018deeprl}, evaluation in the few-seed regime must report
uncertainty rather than a point estimate~\citep{agarwal2021precipice}, and the
community has codified reporting practices to combat this
failure~\citep{pineau2021reproducibility}. An
agent that experiments faster only compounds the over-claiming risk unless its
conclusions are filtered. Every claim is therefore denominated in a measured unit
of noise, and the three checks that follow are applied uniformly to every family.

\paragraph{Measured seed noise.}
Before any improvement is credited, the framework measures how much the metric moves from one
random seed to the next, for reasons unrelated to the agent's edits. This is
the family's \emph{seed noise}: the run-to-run variability a real gain must exceed,
analogous to the noise floor a signal must rise above to be detected. A fixed baseline configuration
$\theta_0$ is run across $K$ seeds, yielding
\begin{equation}
\mu_0 \;=\; \frac{1}{K}\sum_{i=1}^{K} f(\theta_0; s_i),
\qquad
\sigma \;=\; \sqrt{\frac{1}{K-1}\sum_{i=1}^{K}\bigl(f(\theta_0; s_i)-\mu_0\bigr)^2},
\qquad K\ge 2,
\label{eq:noisefloor}
\end{equation}
the sample mean and the Bessel-corrected sample standard deviation of the metric.
The quantity $\sigma$ is not a tuned hyperparameter but a measured property of the
family, and it is the unit in which all subsequent claims are denominated. Because
$\sigma$ is itself estimated from finitely many seeds, $K$ must be large enough
that the acceptance margin below is not dominated by uncertainty in $\sigma$.
$K$ is therefore reported with every result. In the demonstration of
Sec.~\ref{sec:results} the mean difference and its confidence interval are also
reported, alongside the standardized margin, so the claim does not rest on a
sigma-count drawn from too few samples. This is the discipline the few-seed
literature calls for~\citep{agarwal2021precipice}, applied to the noise estimate
itself.

The same $\sigma$ also supplies the in-loop machinery of Sec.~\ref{sec:loop}. After
each run the orchestrator assigns a cheap, single-seed verdict relative to the
prior best $b_{t-1}$,
\begin{equation}
v(\theta_t) \;=\;
\begin{cases}
\text{confirmed} & \delta \ge \hat t_{\mathrm{nf}},\\[2pt]
\text{regressed} & \delta \le -\hat t_{\mathrm{nf}},\\[2pt]
\text{within noise} & |\delta| < \hat t_{\mathrm{nf}},
\end{cases}
\qquad
\delta = \sigma_{\mathrm{dir}}\bigl(f(\theta_t;s_t)-b_{t-1}\bigr),
\label{eq:perrun}
\end{equation}
where the threshold $\hat t_{\mathrm{nf}}$ is the measured $\sigma$ when available
and a fallback of $\max\!\bigl(0.05\,|b_{t-1}|,\,10^{-12}\bigr)$ otherwise. The same
threshold drives the plateau predicate that feeds the stopping time of
Eq.~\eqref{eq:stop}: with a sliding window of the last $w$ successful runs (here
$w=6$) and the pre-window best $b^{\mathrm{pre}}$,
\begin{equation}
\Pi_t \;=\;
\Bigl[\;\forall\, r \in \text{last } w \text{ runs}:\;
\sigma_{\mathrm{dir}}\bigl(f_r - b^{\mathrm{pre}}\bigr) \;<\; \kappa\,\hat t_{\mathrm{nf}}\;\Bigr],
\label{eq:plateau}
\end{equation}
so the campaign is declared stuck when an entire window fails to beat the
pre-window best by more than $\kappa$ noise units (with scale $\kappa=1$ by
default).

\paragraph{Reseeded verification.}
A single-seed best run can be a fluke of the seed, so the agent's reported best is
not believed on the strength of the run that produced it. The layer reconstructs each
configuration's full setting by walking \code{proposals.jsonl}, deduplicates by
configuration with the seed removed, ranks the candidates, and re-runs the top
few at a set of fresh, held-out seeds disjoint from those that generated them. A
train-loss parity guard excludes any reseed whose best training loss exceeds
twice that of its source run, so a seed throttled by resource contention cannot
pollute the aggregate. Over the valid reseeds $\mathcal{S}_R$ the
reseeded mean $\mu_R$ and standard deviation $\sigma_R$ are formed, and $\mu_R$, not
the single-seed best, is reported as the headline number. The improvement is credited only
when it clears the seed noise by a standardized margin and is feasible on every
seed,
\begin{equation}
\text{accept }\hat\theta
\iff
\underbrace{\sigma_{\mathrm{dir}}\bigl(\mu_R-\mu_0\bigr)\;\ge\;2\sigma}_{\text{effect-size gate}}
\;\;\wedge\;\;
\underbrace{\textstyle\bigwedge_{s\in\mathcal{S}_R}\mathrm{Feasible}(\hat\theta;s)}_{\text{all-seeds safety}},
\qquad
m \;=\; \frac{\sigma_{\mathrm{dir}}\bigl(\mu_R-\mu_0\bigr)}{\sigma}.
\label{eq:reseedgate}
\end{equation}
The gate is a one-sided test in units of the measured standard deviation, in the
spirit of a standardized effect size~\citep{cohen1988statistical}. The reported
margin $m$ counts how many noise units separate the reseeded result from the
baseline, and a result is promoted to the family best only when $m\ge 2$ and the
hard safety constraint holds on all of $\mathcal{S}_R$. For the rendezvous
campaign of Sec.~\ref{sec:results} this margin is roughly
$m\approx\cwhSigmaMargin$.

\paragraph{Leave-one-out pruning.}
Acceptance establishes that the result is real; pruning asks which of the agent's
edits actually carry it. Starting from the accepted configuration $\hat\theta$,
the layer reverts each modified parameter $i$ to its family default in isolation, holds the
rest fixed, and remeasures. Writing the oriented change from that reversion as
\begin{equation}
\Delta_i \;=\; \sigma_{\mathrm{dir}}\bigl(f(\theta_{-i}) - \mu_R\bigr),
\qquad
\theta_{-i} = \hat\theta \text{ with parameter } i \text{ reset to default,}
\label{eq:loo}
\end{equation}
a negative $\Delta_i$ means reverting the parameter hurts, so the parameter is
load-bearing, while a positive $\Delta_i$ means reverting it helps, so the agent
moved it the wrong way. The decision is taken in noise units and escalated only
when a single probe seed is inconclusive, with $\Delta_i^{(1)}$ the one-seed
estimate and $\Delta_i^{(2)}$ a two-seed mean,
\begin{equation}
\mathrm{decision}(i) \;=\;
\begin{cases}
\text{keep} & \Delta_i^{(1)} \le -\sigma \quad\text{(clearly load-bearing),}\\[4pt]
\text{drop} & \bigl|\Delta_i^{(2)}\bigr| < 0.5\,\sigma \quad\text{(no signal, a hitchhiker),}\\[4pt]
\text{drop}^{\dagger} & \Delta_i^{(2)} > 0 \quad\text{(reverting helps, flag interaction),}\\[4pt]
\text{keep} & -\sigma < \Delta_i^{(2)} \le -0.5\,\sigma \quad\text{(borderline, kept conservatively).}
\end{cases}
\label{eq:loodecision}
\end{equation}
This separates load-bearing edits from inert hitchhikers and, through the
$\dagger$ case, surfaces edits moved in the wrong direction. A subtlety the
single-parameter test cannot see is that a group of parameters can each be individually
inert, $|\Delta_i^{(2)}|<0.5\sigma$, yet matter jointly, so that dropping all of
them together regresses the metric. The layer guards against this by re-verifying
the pruned recipe $\theta_{\mathrm{pruned}}$ at fresh seeds and retaining the
simplification only if it still clears the same gate,
\begin{equation}
\text{publish }\theta_{\mathrm{pruned}}
\iff
\sigma_{\mathrm{dir}}\bigl(F(\theta_{\mathrm{pruned}})-\mu_0\bigr) \ge 2\sigma
\;\;\wedge\;\;
\text{target retained;}
\label{eq:prunereverify}
\end{equation}
otherwise the full configuration is kept. In the rendezvous campaign this is what
happens: five edits prove load-bearing and four are individually inert, but the
four together are not, so the re-verification of Eq.~\eqref{eq:prunereverify}
forces the full recipe to be retained.

Finally, as an external sanity check and not as a promotion gate, the framework runs the
undirected random search of Eq.~\eqref{eq:randsearch} over the same editable
surface from the same shared first iteration under a matched per-experiment
budget. Comparing the two arms' best results in the same noise units gives a
verdict,
\begin{equation}
V \;=\;
\begin{cases}
\text{LLM wins} & \sigma_{\mathrm{dir}}\bigl(b^{\mathrm{LLM}}-b^{\mathrm{rs}}\bigr) > 2\sigma,\\[2pt]
\text{random wins} & \sigma_{\mathrm{dir}}\bigl(b^{\mathrm{LLM}}-b^{\mathrm{rs}}\bigr) < -2\sigma,\\[2pt]
\text{tie} & \text{otherwise,}
\end{cases}
\label{eq:verdict}
\end{equation}
which confirms that the agent's guidance does more than blind sampling
would, while keeping the seed-noise gate of Eq.~\eqref{eq:reseedgate} as the
claim on which the result actually rests.

%% file: sections/fig_loop.tex
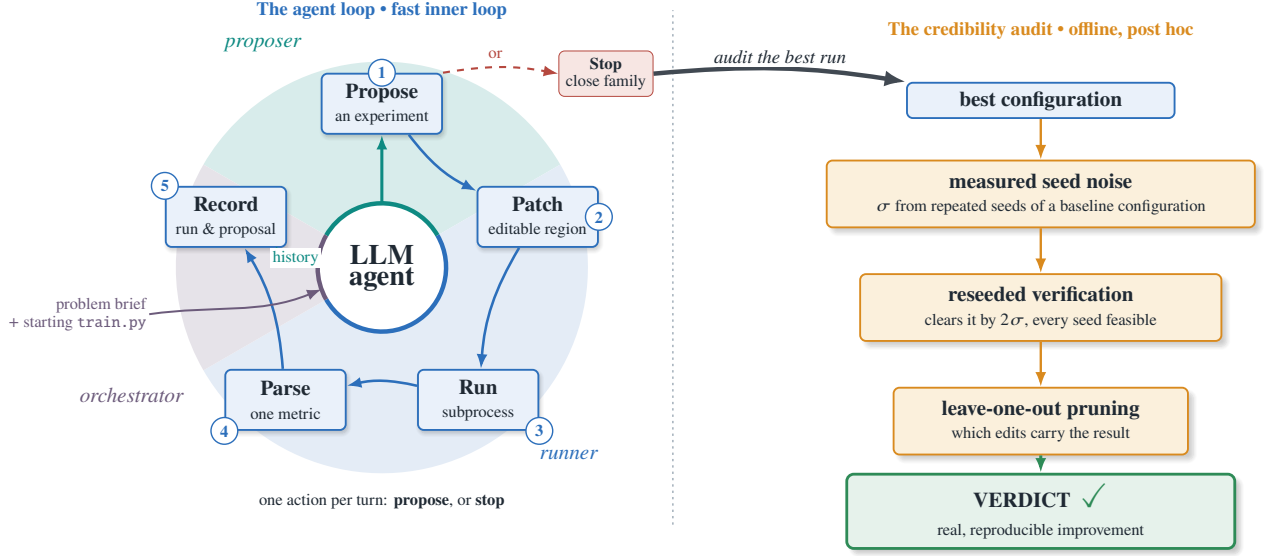
\begin{figure}[t]
\centering
\resizebox{\linewidth}{!}{%
\begin{tikzpicture}[
    font=\small,
    >={Latex[length=2.2mm]},
    stage/.style={draw=cStage, line width=0.8pt, rounded corners=2.5pt, fill=cStageBg,
                  align=center, inner sep=3.5pt, text=ink, minimum width=18mm, minimum height=9mm,
                  blur shadow={shadow blur steps=6, shadow xshift=0.25mm,
                               shadow yshift=-0.45mm, shadow opacity=24}},
    flow/.style={->, line width=1.1pt, draw=cStage},
    badge/.style={circle, draw=cStage, fill=white, text=cStage, line width=0.6pt, inner sep=0pt,
                  minimum size=4.0mm, font=\scriptsize\bfseries},
    filt/.style={draw=cCred, line width=0.8pt, rounded corners=2.5pt, fill=cCredBg, align=center,
                 inner sep=4pt, text=ink, minimum height=10mm,
                 blur shadow={shadow blur steps=6, shadow opacity=20}},
]
\def\Rr{2.45}
\coordinate (O) at (-5.4,0);

\begin{scope}[on background layer]
  \fill[cAgent!10] (O) circle (12.5mm);
\end{scope}

\fill[cAgent!16] (O) -- ($(O)+(30:3.08)$)   arc[start angle=30,   end angle=150, radius=3.08] -- cycle;
\fill[cStage!13] (O) -- ($(O)+(-150:3.08)$)  arc[start angle=-150, end angle=30,  radius=3.08] -- cycle;
\fill[cOrch!17]  (O) -- ($(O)+(150:3.08)$)   arc[start angle=150,  end angle=210, radius=3.08] -- cycle;

\node[font=\footnotesize\bfseries, text=cStage] at ($(O)+(0,3.85)$)
      {The agent loop \textbullet{} fast inner loop};
\def\Ra{0.96cm}
\fill[white] (O) circle (\Ra);
\draw[cAgent, line width=1.9pt] ($(O)+(30:\Ra)$)  arc[start angle=30,  end angle=150, radius=\Ra];
\draw[cOrch,  line width=1.9pt] ($(O)+(150:\Ra)$) arc[start angle=150, end angle=210, radius=\Ra];
\draw[cStage, line width=1.9pt] ($(O)+(210:\Ra)$) arc[start angle=210, end angle=390, radius=\Ra];
\node[circle, text=ink, align=center, inner sep=1pt, minimum size=19mm] (agent) at (O)
      {\textbf{\large LLM}\\[-2pt]\textbf{\large agent}};

\node[stage] (prop)  at ($(O)+(90:\Rr)$)   {\textbf{Propose}\\[-1pt]\scriptsize an experiment};
\node[stage] (patch) at ($(O)+(18:\Rr)$)   {\textbf{Patch}\\[-1pt]\scriptsize editable region};
\node[stage] (run)   at ($(O)+(-54:\Rr)$)  {\textbf{Run}\\[-1pt]\scriptsize subprocess};
\node[stage] (parse) at ($(O)+(-126:\Rr)$) {\textbf{Parse}\\[-1pt]\scriptsize one metric};
\node[stage] (rec)   at ($(O)+(162:\Rr)$)  {\textbf{Record}\\[-1pt]\scriptsize run \& proposal};

\node[badge] at (prop.north)       {1};
\node[badge] at (patch.east)       {2};
\node[badge] at (run.south east)   {3};
\node[badge] at (parse.south west) {4};
\node[badge] at (rec.north west)   {5};

\draw[flow] (prop)  to[bend right=13] (patch);
\draw[flow] (patch) to[bend right=13] (run);
\draw[flow] (run)   to[bend right=13] (parse);
\draw[flow] (parse) to[bend right=13] (rec);

\draw[->, line width=1.3pt, draw=cAgent] (agent) -- (prop);
\draw[->, line width=1.0pt, draw=cAgent, dashed] (rec) to[bend right=14]
      node[pos=0.5, fill=white, inner sep=1pt, font=\scriptsize, text=cAgent]{history} (agent);

\node[font=\scriptsize, text=ink, align=center] at ($(O)+(0,-3.5)$)
      {one action per turn: \textbf{propose}, or \textbf{stop}};

\node[font=\scriptsize, text=cOrch, align=right, inner sep=1pt] (inputs) at (-9.95,-0.7)
      {problem brief\\$+$ starting \texttt{train.py}};
\draw[->, line width=0.85pt, draw=cOrch] (inputs.east) to[out=8,in=202] (agent.200);

\node[rounded corners=2.5pt, draw=cStop, fill=cStopBg, text=ink, align=center,
      inner sep=3pt, font=\scriptsize] (stop) at (-2.05,2.9)
      {\textbf{Stop}\\[-1pt]close family};
\draw[->, line width=0.9pt, draw=cStop, dashed] (prop.north east) to[bend left=10]
      node[pos=0.45, above, font=\scriptsize, text=cStop]{or} (stop.west);

\node[font=\small\itshape, text=cAgent] at ($(O)+(118:3.78)$) {proposer};
\node[font=\small\itshape, text=cStage] at ($(O)+(-45:3.95)$) {runner};
\node[font=\small\itshape, text=cOrch]  at ($(O)+(207:4.2)$) {orchestrator};

\draw[orbit, line width=0.7pt, dash pattern=on 1pt off 1.7pt] (-1.05,-3.85) -- (-1.05,3.85);

\def\cx{4.45}
\node[font=\footnotesize\bfseries, text=cCred] at (\cx,3.55)
      {The credibility audit \textbullet{} offline, post hoc};

\node[draw=cStage, line width=0.8pt, rounded corners=2.5pt, fill=cStageBg, text=ink,
      align=center, inner sep=3.5pt, minimum width=40mm] (best) at (\cx,2.5)
      {\textbf{best configuration}};

\node[filt, minimum width=64mm] (f1) at (\cx,1.1)
      {\textbf{measured seed noise}\\[-1pt]\scriptsize $\sigma$ from repeated seeds of a baseline configuration};
\node[filt, minimum width=54mm] (f2) at (\cx,-0.6)
      {\textbf{reseeded verification}\\[-1pt]\scriptsize clears it by $2\sigma$, every seed feasible};
\node[filt, minimum width=44mm] (f3) at (\cx,-2.3)
      {\textbf{leave-one-out pruning}\\[-1pt]\scriptsize which edits carry the result};

\draw[->, line width=1.0pt, draw=cCred] (best) -- (f1);
\draw[->, line width=1.0pt, draw=cCred] (f1) -- (f2);
\draw[->, line width=1.0pt, draw=cCred] (f2) -- (f3);

\node[draw=cWin, line width=1.2pt, rounded corners=3pt, fill=cWinBg, text=ink, align=center,
      inner sep=4pt, minimum width=58mm, font=\small,
      blur shadow={shadow blur steps=6, shadow opacity=22}] (verdict) at (\cx,-3.65)
      {\textbf{VERDICT}\ \ \textcolor{cWin}{\Large\checkmark}\\[1pt]
       \scriptsize real, reproducible improvement};
\draw[->, line width=1.3pt, draw=cWin] (f3) -- (verdict);

\draw[-{Latex[length=3.2mm]}, line width=1.9pt, draw=ink!85] (stop.east) to[bend left=7]
      node[pos=0.5, above, font=\footnotesize\itshape, text=ink, fill=white, inner sep=1.5pt]
      {audit the best run} (best.north west);
\end{tikzpicture}%
}
\caption{The AutoResearch agent loop and its credibility audit. \emph{Left:} the
language model drives a fast inner loop, reading
the problem description, the run history, and the current hyperparameters and emitting
one action per turn, either an experiment or a stop. The
three tinted zones on the loop mark its proposer, runner, and orchestrator modules
(Sec.~\ref{sec:loop}). A proposed experiment travels once around the loop: the runner
patches the editable region of \code{train.py}, runs it, parses one metric line, logs
it, and feeds the updated history back. \emph{Right:} when the loop stops, its best
configuration enters an offline credibility audit that measures a per-problem seed
noise, reseeds the candidate to confirm it clears it by a fixed margin under
fresh seeds, and prunes its individual edits. The audit, not the raw loop, is the
step that confirms a genuine improvement reproduces across fresh seeds.}
\label{fig:loop}
\end{figure}

%% file: sections/04_problems.tex
\section{Demonstration Problems}
\label{sec:problems}

The framework is exercised on two spacecraft-guidance problems that share the same
Clohessy--Wiltshire relative dynamics and pose the same research task: learn, by
behavioral cloning, a feedback policy that matches an optimal-control solver across a
wide range of initial conditions. They differ in just one way that matters for the
audit. The first is an unconstrained \emph{rendezvous}, in which the chaser may fly any
path to the target. The second is a \emph{collision-avoidance docking}, in which the
chaser must reach a point just outside a spherical keep-out zone while starting on its
far side, so the straight-line approach pierces the zone and the policy must actively
detour around it. The second therefore binds a hard safety constraint that the first
does not.

These problems are chosen not for their guidance content, which is classical, but because
each is \emph{bracketed} in a way that lets the credibility layer
(Sec.~\ref{sec:credibility}) do its work. At the bottom, the optimal-control expert
defines a known, near-zero performance benchmark; just above it, a fixed baseline
configuration fixes the scale of seed-to-seed metric noise. An apparent improvement can
then be judged against both a meaningful benchmark and a measured noise, which is what
the layer needs to certify that a gain is real and not just a lucky seed. The docking
problem adds a binding safety constraint that an accuracy-only metric cannot capture,
stressing the same layer along a second axis. In both, the contribution is the
autonomous, auditable process that produces the policy, not a new guidance law.

\subsection{Relative orbital dynamics}
\label{sec:dynamics}
The chaser's motion relative to the target is modeled with the linear
Clohessy--Wiltshire--Hill (CWH) equations~\citep{clohessy1960terminal} in the
target's local-vertical/local-horizontal (LVLH) frame. Writing the relative
position as $\mathbf{r}=[x,y,z]^\top$, the per-axis thrust as
$\mathbf{u}=[u_x,u_y,u_z]^\top$, the deputy mass as $m$, and the chief mean motion
as $n$, the continuous-time equations of motion are
\begin{equation}
\label{eq:cwh_eom}
\ddot{x} - 2n\dot{y} - 3n^2 x = \frac{u_x}{m},\qquad
\ddot{y} + 2n\dot{x} = \frac{u_y}{m},\qquad
\ddot{z} + n^2 z = \frac{u_z}{m}.
\end{equation}
With the state $\mathbf{x}=[\mathbf{r}^\top,\dot{\mathbf{r}}^\top]^\top\in\mathbb{R}^6$,
these collect into the linear state-space form
$\dot{\mathbf{x}} = A_c\mathbf{x} + B_c\mathbf{u}$, with
\begin{equation}
\label{eq:cwh_ss}
A_c = \begin{bmatrix} 0_3 & I_3 \\[2pt] \Gamma & \Omega \end{bmatrix},\quad
\Gamma = \begin{bmatrix} 3n^2 & 0 & 0 \\ 0 & 0 & 0 \\ 0 & 0 & -n^2 \end{bmatrix},\quad
\Omega = \begin{bmatrix} 0 & 2n & 0 \\ -2n & 0 & 0 \\ 0 & 0 & 0 \end{bmatrix},\quad
B_c = \begin{bmatrix} 0_3 \\[2pt] \tfrac{1}{m} I_3 \end{bmatrix},
\end{equation}
where $I_3$ and $0_3$ are the $3\times3$ identity and zero blocks. The simulator
advances the state with a forward-Euler discretization at step $\Delta t$,
\begin{equation}
\label{eq:cwh_disc}
\mathbf{x}_{k+1} = A\,\mathbf{x}_k + B\,\mathbf{u}_k,\qquad
A = I_6 + \Delta t\,A_c,\qquad B = \Delta t\,B_c,
\end{equation}
and the cloned policy is evaluated in closed loop on these same matrices, so the
student imitates trajectories the simulator can replay exactly. Both problems use
mean motion $n=\SI{1.13e-3}{\radian\per\second}$ (a low-Earth orbit), step
$\Delta t=\SI{10}{\second}$, and deputy mass $m=\SI{500}{\kilogram}$; they differ in
horizon, thrust authority, and constraints, as detailed below.

\subsection{Rendezvous}
\label{sec:problem_rdzv}
The rendezvous problem runs over a horizon of $H=80$ steps (\SI{800}{\second}), with
each thrust axis box-constrained to $\lvert u_i\rvert\le\SI{0.25}{\newton}$. The
chaser must close from a wide range of initial conditions to the target at the
origin.

\paragraph{Expert demonstrations.} The demonstrations come from the \emph{optimal control}: among all thrust histories
that satisfy the linear dynamics, the per-axis thrust box, and a terminal-equality
constraint driving the state to the origin, it is the one of least control effort, the
finite-horizon minimum-energy solution
\begin{equation}
\label{eq:expert_rdzv}
\min_{\{\mathbf{u}_k\}}\ \sum_{k=0}^{H-1}\lVert\mathbf{u}_k\rVert_2^2
\quad\text{s.t.}\quad
\mathbf{x}_{k+1}=A\mathbf{x}_k+B\mathbf{u}_k,\;\;
\mathbf{x}_0=\mathbf{x}^{(i)},\;\;
\mathbf{x}_H=\mathbf{0},\;\;
\lVert\mathbf{u}_k\rVert_\infty\le u_{\max}.
\end{equation}
Because the dynamics are linear and the cost convex, this optimal-control problem is a
convex quadratic program, solved to global optimality. Initial conditions are sampled
from a fixed box (positions in
$[\pm75,\pm75,\pm37.5]\,$m, velocities in $[\pm0.08,\pm0.08,\pm0.04]\,$m/s) and then
\emph{screened for reachability}: Eq.~\eqref{eq:expert_rdzv} is solved for each
candidate, and only initial conditions the expert can drive to the origin within
tolerance are retained. As a result the expert's success rate on the evaluation set
is $1.0$ by construction, which makes the success criterion (below) a genuine test
of the cloned policy's quality rather than an artifact of infeasible starts.

\paragraph{Policy and metric.} The cloned policy is a time-conditioned
multilayer perceptron whose input is the seven-vector
$[x,y,z,\dot x,\dot y,\dot z,\tau]$, where the time-to-go feature $\tau=(H-k)/H$
decreases from $1$ at the first step to $1/H$ at the last. The time feature is
essential: the finite-horizon expert is a time-varying controller, and a policy that
observes only the physical state is asked to imitate a one-to-many map and fails near
the terminal time. Writing $\mathbf{s}$ for this time-augmented policy input, in both problems the
policy $\pi_\theta$ is trained by behavioral cloning on the expert demonstrations:
the input is standardized componentwise to zero mean and unit variance,
$\tilde{\mathbf{s}}=(\mathbf{s}-\boldsymbol{\mu}_{\mathbf{s}})/\boldsymbol{\sigma}_{\mathbf{s}}$,
the expert action is normalized by the thrust limit,
$\tilde{\mathbf{u}}^{\mathrm{e}}=\mathbf{u}^{\mathrm{e}}/u_{\max}$, and the weights
$\theta$ minimize the mean-squared action error
\begin{equation}
\label{eq:bc_loss}
\min_{\theta}\ \frac{1}{N}\sum_{i=1}^{N}
\bigl\lVert\,\pi_\theta(\tilde{\mathbf{s}}_i)-\tilde{\mathbf{u}}^{\mathrm{e}}_i\,\bigr\rVert_2^2
\end{equation}
over the $N$ state-action pairs in the demonstration set (the chosen number of
expert trajectories unrolled over the horizon). At evaluation the policy output is
rescaled and saturated to the thrust box,
$\mathbf{u}=\operatorname{clip}\!\bigl(u_{\max}\,\pi_\theta(\tilde{\mathbf{s}}),\,\pm u_{\max}\bigr)$,
and applied in closed loop. The primary metric is the \emph{mean terminal
distance}, the mean Euclidean position error at the final step over a held-out set
of $128$ reachability-screened initial conditions, in meters, with lower being
better. A per-condition success requires a
terminal distance below \SI{25}{\meter} and a terminal speed below
\SI{0.05}{\meter\per\second}. The
optimal-control expert reaches the origin essentially exactly, so it sets a near-zero
performance benchmark; the learning question is how close a behaviorally cloned policy can
come to it.

\subsection{Collision-avoidance docking}
\label{sec:problem_rpo}
The docking problem uses the same dynamics (Eqs.~\eqref{eq:cwh_eom}--\eqref{eq:cwh_disc})
over a longer horizon of $H=240$ steps (\SI{2400}{\second}) and with a larger per-axis
thrust box $\lvert u_i\rvert\le\SI{2}{\newton}$, to give the detour the needed control
authority. The chief occupies a hard keep-out sphere of radius
$R_{\mathrm{KOZ}}=\SI{5}{\meter}$ centered at the LVLH origin $\mathbf{c}$. The docking
target $\mathbf{g}$ sits at $R_{\mathrm{goal}}=\SI{8}{\meter}$, \SI{3}{\meter} outside
the keep-out surface, on a per-episode random bearing, and the chaser starts
\SIrange{40}{110}{\meter} out on the far side, biased opposite the docking point, so a
direct run to the target would cross the sphere.

\paragraph{Expert demonstrations.} The expert is again the finite-horizon minimum-energy optimal control,
driving both the position and the relative velocity to the docking state
while enforcing the keep-out as a sequence of per-step radial half-space constraints, in
the spirit of convex-optimization guidance~\citep{acikmese2007convex},
\begin{equation}
\label{eq:expert_rpo}
\min_{\{\mathbf{u}_k\}}\ \sum_{k=0}^{H-1}\lVert\mathbf{u}_k\rVert_2^2
\quad\text{s.t.}\quad
\begin{aligned}[t]
&\mathbf{x}_{k+1}=A\mathbf{x}_k+B\mathbf{u}_k,\quad
\mathbf{x}_0=\mathbf{x}^{(i)},\quad
\mathbf{x}_H=[\mathbf{g}^\top,\mathbf{0}^\top]^\top,\\
&\lVert\mathbf{u}_k\rVert_\infty\le u_{\max},\qquad
\hat{\mathbf{n}}_k^\top(\mathbf{r}_k-\mathbf{c})\ge R_{\mathrm{KOZ}}+\delta,
\end{aligned}
\end{equation}
where $\hat{\mathbf{n}}_k$ is the unit radial direction of the current iterate's
position from the keep-out center and $\delta$ is a standoff margin. The half-space is a
convex restriction of the non-convex ``stay outside the ball'' constraint, refined over
a few sequential-convex-programming iterations into a smooth detour that hugs the
keep-out and ends exactly at the goal. The resulting trajectories are collision-free and
terminal-zeroing, so the expert defines a near-zero accuracy benchmark and is
feasible by construction. As in the first family the expert is time-varying, so the
cloned policy is given a time-to-go feature, and only collision-free, goal-reaching
demonstrations survive a reachability screen.

\paragraph{Policy, metric, and safety gate.} The cloned policy is a time-conditioned
multilayer perceptron mapping the ten-vector
$[\,\mathbf{r}^\top,\mathbf{v}^\top,\mathbf{g}^\top,\tau\,]^\top$ (relative
position, relative velocity, the docking-target position $\mathbf{g}$, and the
normalized time-to-go $\tau$) to a three-axis thrust command, and it is trained by the
same behavioral-cloning objective as the rendezvous policy (Eq.~\eqref{eq:bc_loss}),
differing only in its inputs and in the metric below. Evaluation is pure closed-loop
rollout, never anchored to the expert reference, so compounding error is visible by
design. The primary metric is a safety-aware docking score, in
meters-equivalent and lower is better,
\begin{equation}
\textsc{score} \;=\; \bar d_{\mathrm{T}} \;+\; 10\,\bar\nu,
\label{eq:rpo_score}
\end{equation}
where $\bar d_{\mathrm{T}}$ is the mean terminal distance to the docking point and
$\bar\nu$ is the mean fraction of rollout steps spent inside the keep-out zone, both
averaged over a held-out set of $128$ far-side initial conditions. The
$10\times$ weight makes a single sustained violation cost more than the entire
docking-accuracy budget, so accuracy cannot be bought with violations. Feasibility itself
is a separate hard gate rather than a score term: a \emph{strict} run requires a success
rate of at least $0.80$ (an initial condition succeeds when the terminal distance and
speed are within tolerance with zero keep-out violations) \emph{and} a mean violation
rate of exactly zero \emph{and} a nonnegative minimum keep-out clearance. A run is
credited only if it lowers Eq.~\eqref{eq:rpo_score} \emph{and} clears this gate; the
family target is a score at or below $0.7$ under the gate.

\paragraph{Run-time safety lever.} Because the constraint is hard, the editable surface
includes a discrete-time predictive control-barrier-function (CBF) safety
filter~\citep{ames2019cbf} that can minimally edit each thrust command at rollout time to
keep the chaser outside the keep-out zone, the standard run-time-assurance paradigm. For
this relative-degree-two system a one-step lookahead cannot be filtered, so the filter
predicts the two-step state under a held action,
$\mathbf{x}_{t+2}=A^2\mathbf{x}_t+(AB+B)\mathbf{u}$, and requires the radial component of
the predicted position to stay outside $R_{\mathrm{KOZ}}+\delta$. This reduces to a
single-constraint quadratic program that minimally corrects the base-policy command
$\mathbf{u}_{\mathrm{BC}}$,
\begin{equation}
\label{eq:cbf_qp}
\mathbf{u}^\star=\arg\min_{\mathbf{u}}\ \lVert\mathbf{u}-\mathbf{u}_{\mathrm{BC}}\rVert_2^2
\quad\text{s.t.}\quad
\boldsymbol{\beta}^\top\mathbf{u}\ge\rho,\qquad
\lVert\mathbf{u}\rVert_\infty\le u_{\max},
\end{equation}
with $\boldsymbol{\beta}=\big[(AB+B)^\top\hat{\mathbf{n}}\big]_{1:3}$ and
$\rho=(R_{\mathrm{KOZ}}+\delta)-\hat{\mathbf{n}}^\top\big[A^2\mathbf{x}_t\big]_{1:3}$,
where $\hat{\mathbf{n}}$ is the radial direction anchored at $\mathbf{r}_{t+1}$ and
$[\,\cdot\,]_{1:3}$ takes the position rows. It has a closed-form solution (half-space
projection plus box clip) and is exposed to the agent as a single Boolean parameter: with the
filter off, even a well-trained policy clips the keep-out on its boundary-riding detours
and is infeasible; with it on, the keep-out is enforced and the policy is strict-feasible.
The filter is a soft assurance, not a guarantee, since a sufficiently poor policy can
still miss the dock. Whether to engage it is one of the choices the agent must make, and
it lets the credibility layer separate an accuracy improvement from a safety one.

\subsection{Editable surface}
\label{sec:editable}
Across both problems the agent may edit only the hyperparameters in
Table~\ref{tab:editable}; the dynamics, the keep-out geometry, the experts, the datasets,
the evaluation protocol, the penalty weight, the filter standoff $\delta$, and input-noise
augmentation (fixed off at zero in both families) are immutable. In blind mode the agent
is shown the \emph{operational bounds} of each parameter, that is, the values the runner and
dataset accept, but not which values are good; it must discover the good region
empirically. The description also defines an \emph{axis taxonomy} (data, optimizer,
architecture, training budget, regularization, and, for the docking problem, safety) that
the agent must use to classify each proposal, so the recorded trail reveals which parts of
the search space were actually explored.

\begin{table}[!ht]
\centering
\caption{The editable hyperparameter surface shared by the two demonstration problems, as
shown to the agent in blind mode. Bounds are operational limits, not hints about the
optimum, and the code name of each parameter is given for reference. The last two parameters exist
only in the docking problem; ``n/a'' marks a parameter absent from the rendezvous
problem. The random-search baseline samples uniformly from a bounded sub-range of these
same parameters.}
\label{tab:editable}
\small
\begin{tabular}{l l l l l}
\toprule
Hyperparameter & Code name & Axis & Rendezvous & Docking \\
\midrule
Demonstrations   & \code{N\_DEMOS}                & data            & int $[1,256]$        & $\{32,64,128,256\}$ \\
Training epochs  & \code{EPOCHS}                  & training budget & int $[5,5000]$       & int $[10,120]$ \\
Learning rate    & \code{LEARNING\_RATE}          & optimizer       & $[10^{-5},10^{-2}]$  & $[10^{-4},5\times10^{-3}]$ \\
Batch size       & \code{BATCH\_SIZE}             & optimizer       & int $[16,1024]$      & $\{64,128,256,512\}$ \\
Weight decay     & \code{WEIGHT\_DECAY}           & optimizer       & $[0,10^{-2}]$        & $[0,10^{-3}]$ \\
Gradient clip    & \code{GRAD\_CLIP}              & optimizer       & $[0.1,10]$           & $[0.5,5]$ \\
Hidden width     & \code{HIDDEN\_DIM}             & architecture    & int $[16,1024]$      & $\{64,128,256\}$ \\
Depth            & \code{N\_LAYERS}               & architecture    & int $[1,8]$          & int $[1,4]$ \\
Activation       & \code{ACTIVATION}              & architecture    & tanh / ReLU / GELU   & tanh / ReLU / GELU \\
Dropout          & \code{DROPOUT}                 & regularization  & $[0,0.5)$            & $[0,0.3]$ \\
Safety filter    & \code{RUNTIME\_CBF\_FILTER}    & safety          & n/a          & \{off, on\} \\
Recovery augment & \code{RECOVERY\_AUGMENT\_FRAC} & data            & n/a          & $[0,0.5]$ \\
\bottomrule
\end{tabular}
\end{table}

%% file: sections/05_results.tex
\section{Results}
\label{sec:results}

The agent was run on both demonstration problems in blind mode
(Sec.~\ref{sec:proposer}), each for a single campaign, and the identical
credibility check (Sec.~\ref{sec:credibility}) was applied to the best configuration it
found. The outcome takes the same three-part shape on both: the agent drives the cloned
policy down to the numerical neighborhood of the optimal-control benchmark; reseeding on
fresh seeds confirms the gain is real, not a one-seed accident; and a parallel
undirected search over the same parameters trails it. The rendezvous shows the audit in
the clean, unconstrained case; the docking then adds a hard keep-out constraint, so the
audit must certify accuracy and safety at once and the gap to undirected search sharpens
from quantitative to categorical. The two are taken in turn.

\subsection{Rendezvous}
\label{sec:results_rdzv}
\begin{figure}[tbp]
\centering
{\footnotesize\textbf{(a) The agent finds it; undirected search plateaus}}\\[1pt]
\includegraphics[width=0.74\textwidth]{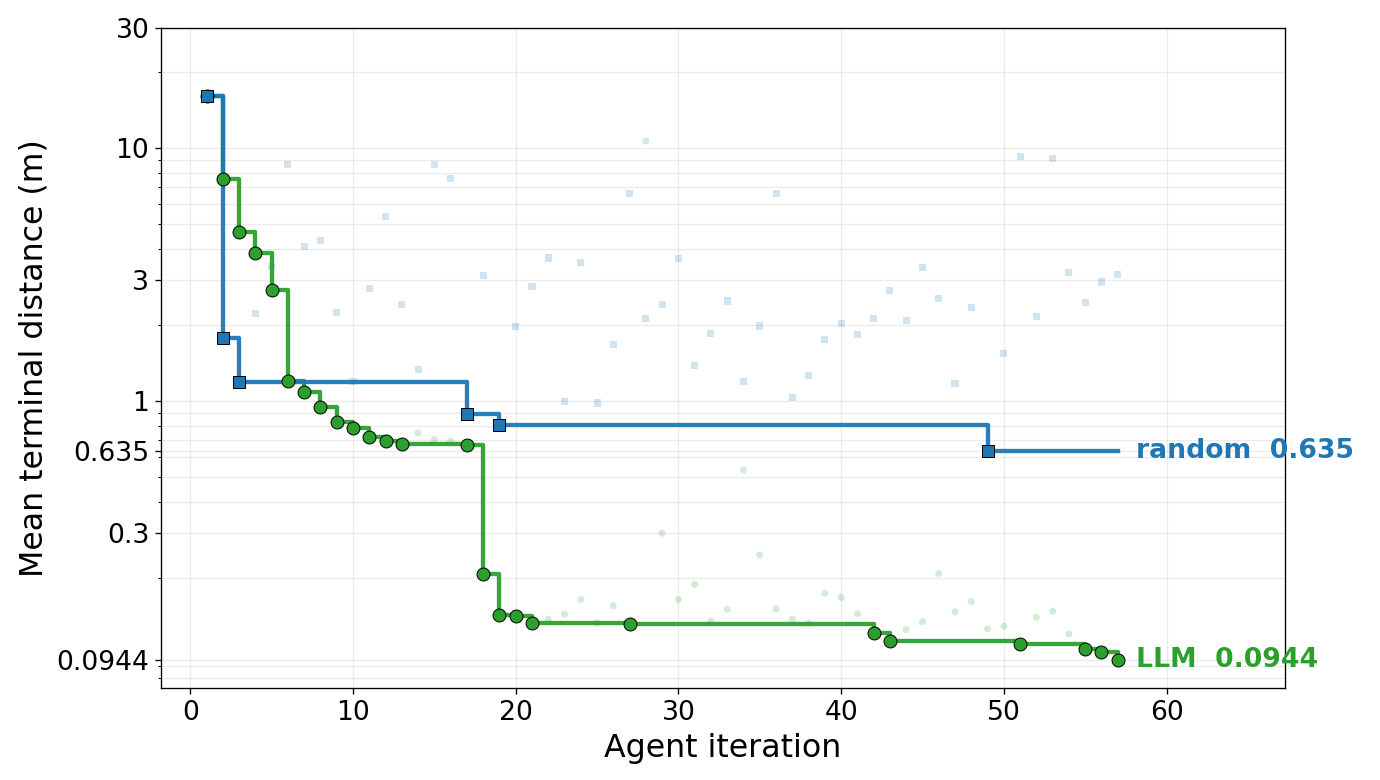}\\[6pt]
{\footnotesize\textbf{(b) \ldots\ and converges on a recipe}}\\[1pt]
\includegraphics[width=0.88\textwidth]{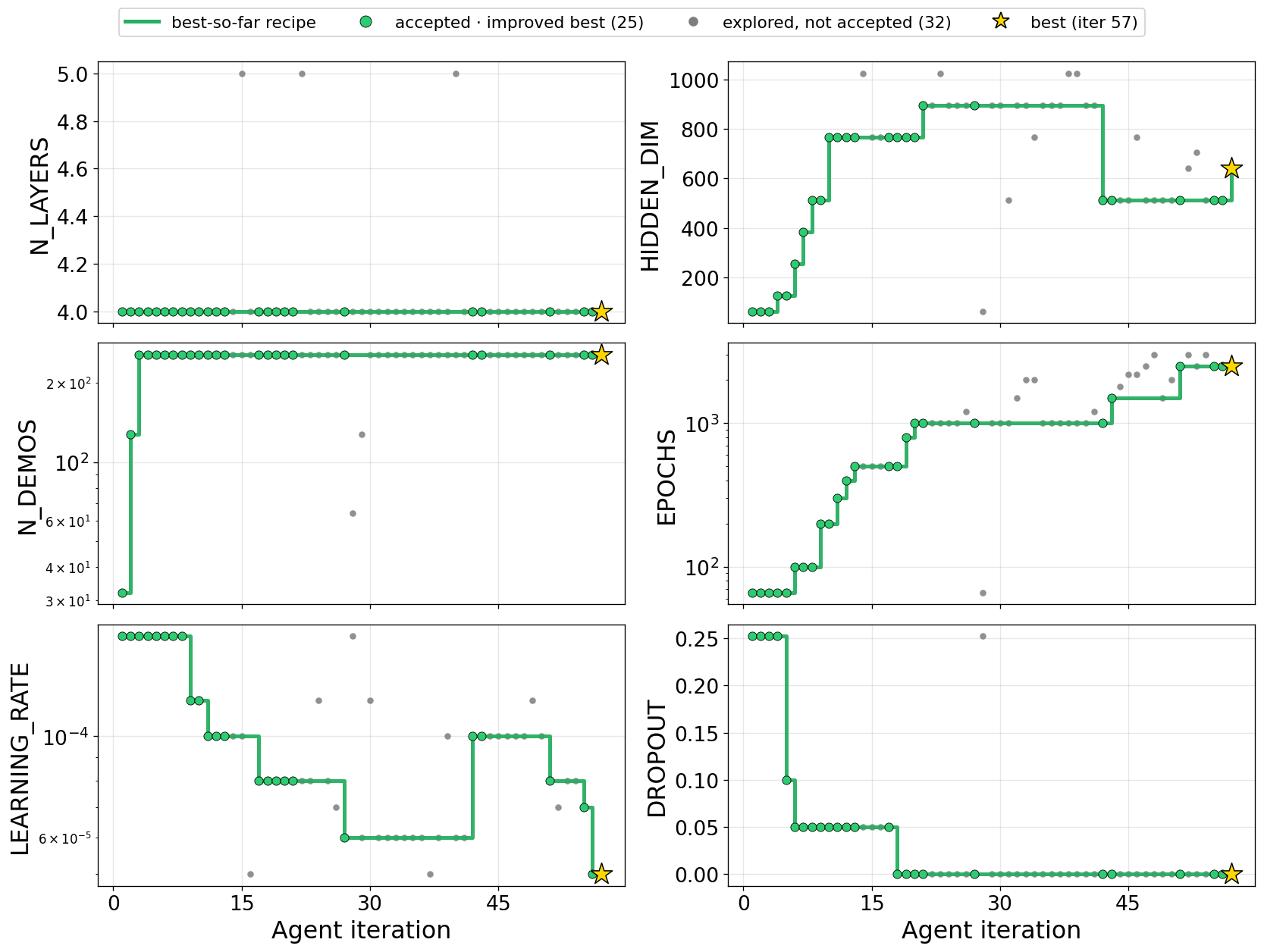}
\caption{The Clohessy--Wiltshire rendezvous search. \textbf{(a)} The agent (green)
drives the cloned policy's mean terminal distance from about \SI{16}{\meter} to
\SI{0.094}{\meter} over \cwhIters{} experiments, while a matched random search (blue)
over the same parameters plateaus near \cwhRand{}. \textbf{(b)} The recipe it converges
on: the network is fixed at four layers early, then the hidden width and training
length grow while the learning rate anneals and dropout falls to zero, a directed
search rather than a random walk.}
\label{fig:hero}
\end{figure}

On the rendezvous problem the agent ran a single blind-mode campaign of \cwhIters{}
experiments. Figure~\ref{fig:hero} summarizes both halves of the result: how the search
descended, and the recipe it converged on.

Both the agent and the random search begin from the same first experiment,
whose policy leaves a mean terminal distance of roughly sixteen meters. From
there the agent works the error down in stages, as the running-best curve in
Fig.~\ref{fig:hero}(a) shows, reaching a single-seed best of \SI{0.094}{\meter}.
That is close to the
optimal-control expert, which reaches the origin to within numerical tolerance. The configuration
behind that single-seed best is wider and trained longer than the default, but no
deeper, holding at four layers: a hidden width of 640 with GELU activations, the
full set of 256 expert demonstrations, a small learning rate, and a large
training budget. Along the way the agent explored still wider networks, up to
roughly 900 units. The trend holds across the campaign (Fig.~\ref{fig:hero}(b)), and it matches the
intuition that a clean imitation problem with an exactly reachable expert rewards
capacity and thorough training over heavy regularization.

\begin{figure}[tbp]
\centering
\includegraphics[width=0.86\textwidth]{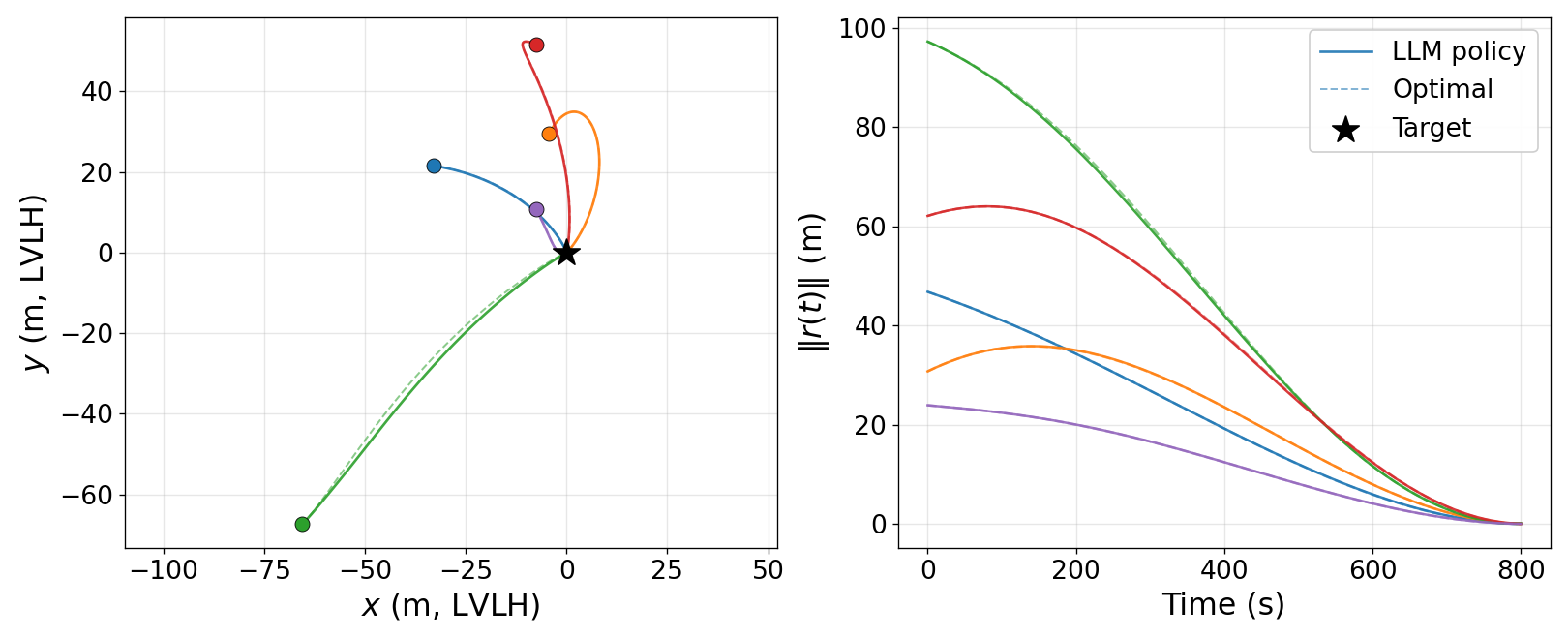}
\caption{The learned policy flies the rendezvous. From five held-out initial
conditions, the cloned policy (solid) steers the spacecraft to the target along
paths that overlay the optimal-control solution (dashed), and the position norms
decay smoothly to zero over the \SI{800}{\second} horizon.}
\label{fig:trajectories}
\end{figure}

Figure~\ref{fig:trajectories} shows what that configuration achieves in the state space
the problem actually cares about. Starting from five held-out initial conditions,
the cloned policy steers the spacecraft to the target along trajectories that
overlay the expert's, and the corresponding position-norm histories decay
smoothly to zero over the eight-hundred-second horizon. The improvement is not a
quirk of the scalar metric; it is a guidance law that flies the rendezvous.

Still, a single-seed best could be luck, and the
credibility check tests that directly. The measured seed noise is
\cwhFloor{}~$\pm$~\cwhFloorSd{} over ten seeds, and at that level the policy succeeds on only about three-quarters of the evaluation cases. Reseeding the agent's best configuration over ten fresh
seeds, disjoint from those that measured it, yields
\cwhLLM{}~$\pm$~\cwhLLMsd{}, with a perfect success rate on every seed. The
reseeded mean improves on the calibrated comparator by a mean difference of
\cwhDelta{} (95\% confidence interval \cwhDeltaCI), which is $m\approx\cwhSigmaMargin$
units of the measured noise standard deviation, far beyond the two-sigma
acceptance threshold. Because that standard deviation is itself measured over ten
seeds, the margin rests on a firm denominator. This gate is the primary evidence that
the result is genuine, resting on nothing more than the problem's own measured noise.

The audit also revises which configuration to designate as best. The campaign's single best
run used one of the wider networks the agent explored, a hidden width of 640, reaching
\SI{0.094}{\meter} at its seed. But under ten fresh seeds a narrower network
(hidden width 512) proved the more reproducible of the two leading candidates, and it
is promoted as the audited family best. The single-seed best (\SI{0.094}{\meter}) and
the reseeded headline (\cwhLLM{}) are therefore different configurations, and keeping
the reproducible one is the whole point of the audit. Table~\ref{tab:cwh} collects the arms.

\begin{table}[t]
\centering
\caption{Credibility-audited results on the Clohessy--Wiltshire rendezvous
problem. Lower mean terminal distance is better; the optimal-control expert reaches the
origin essentially exactly. The reseeded mean is the headline number, and the
single-seed best is shown for traceability.}
\label{tab:cwh}
\small
\begin{tabular}{l c c c}
\toprule
Arm & Mean term.\ dist.\ (m) & Success rate & vs.\ seed noise \\
\midrule
Seed noise (n=10) & $4.04 \pm 0.26$ & $\sim$0.75 & reference \\
Matched random (best of 60)     & $0.635$         & $1.0$      & not gated \\
LLM agent, single-seed best     & $0.094$         & $1.0$      & not gated \\
LLM agent, reseeded (n=10)      & $\mathbf{0.101 \pm 0.007}$ & $\mathbf{1.0}$ & $\mathbf{15.0\sigma}$, \textbf{PASS} \\
\bottomrule
\end{tabular}
\end{table}

As a secondary check, the agent is compared against undirected search over the same
parameters, both anchored to the shared first experiment under a matched per-experiment
budget (Fig.~\ref{fig:hero}(a)). The agent reaches \SI{0.094}{\meter} (reseeded
\cwhLLM{}) while random search across \cwhRandN{} trials plateaus near \cwhRand{}, a
separation of more than six to one. This is best read as suggestive rather than
decisive: the agent reached a larger training budget than the random sweep happened to
sample, so part of the gap reflects the region explored, not the search strategy alone,
and tightening the random arm to a matched range is future work. The central claim
stands on the seed-noise gate above, which does not depend on this comparison.

The credibility check asks not only whether the result is real but which of the
agent's edits produced it. Leave-one-out pruning reverts each edit in turn and
remeasures. Five edits prove load-bearing: the number of demonstrations, the
training budget, the learning rate, the hidden width, and the depth. The remaining
four, batch size, weight decay, gradient clipping, and the activation choice,
carry no individual signal. Dropping all four together, however, regresses
the reseeded metric from \cwhLLM{} to \SI{0.192}{\meter}, so they matter in
combination even though none matters in isolation, and the full configuration is
therefore retained as the family best. This separates edits that matter, those that do
not, and those that matter only in combination, a distinction an informal report would
miss.

\subsection{Collision-avoidance docking}
\label{sec:results_rpo}
\begin{figure}[tbp]
\centering
{\footnotesize\textbf{(a) The agent finds it; undirected search cannot}}\\[1pt]
\includegraphics[width=0.74\textwidth]{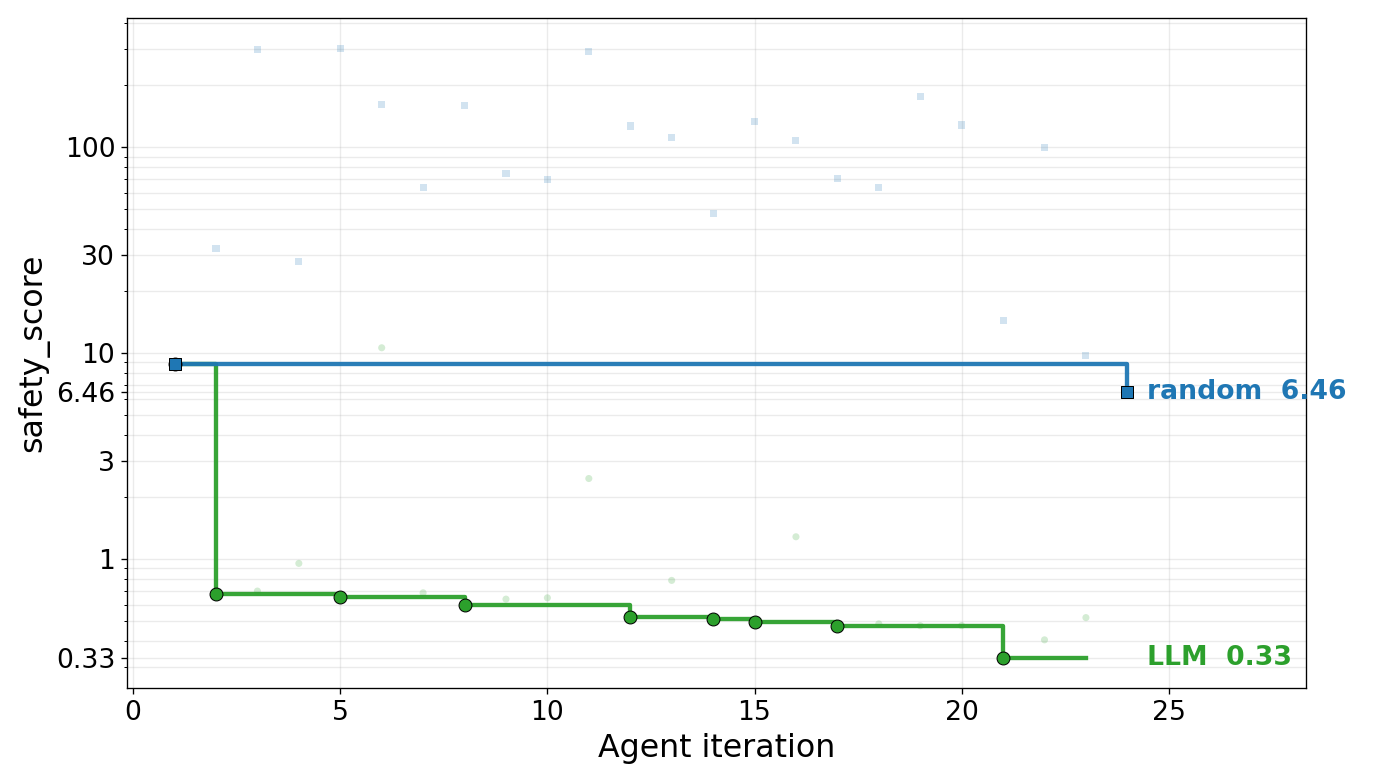}\\[6pt]
{\footnotesize\textbf{(b) \ldots\ and converges on a recipe}}\\[1pt]
\includegraphics[width=0.88\textwidth]{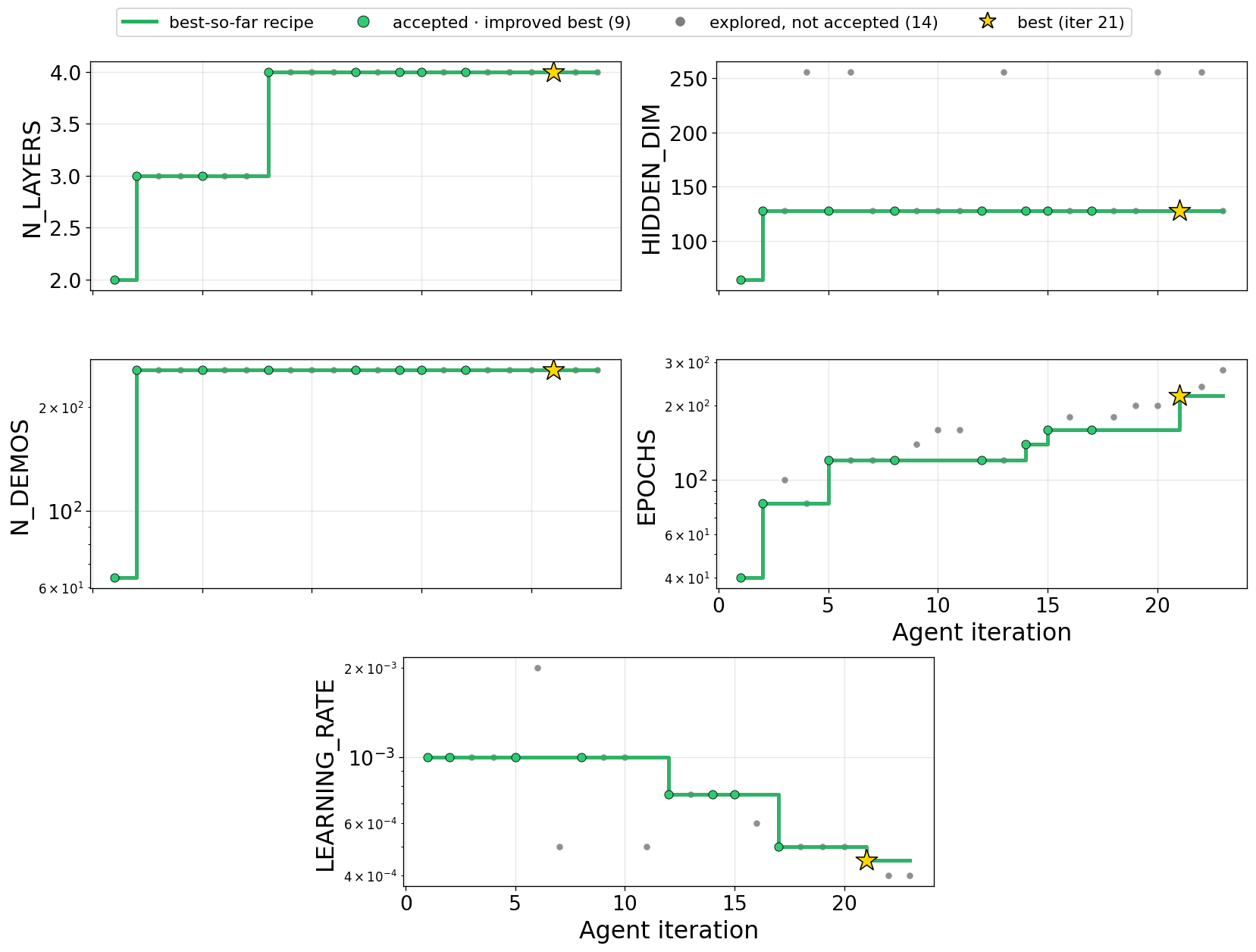}
\caption{The collision-avoidance docking search. \textbf{(a)} The agent (green)
reduces the safety-aware docking score (Eq.~\eqref{eq:rpo_score}, log scale) from its
default to \rpoSingle{} over \rpoIters{} experiments, keeping the safety filter engaged
so every credited run is strict-feasible; a matched random search (blue) never clears
the keep-out gate, reaching only \rpoRand{}. \textbf{(b)} The recipe it converges on:
the full demonstration set early, the network deepened from two to four layers, longer
training, an annealed learning rate, and the run-time CBF filter kept on throughout.
Pruning finds only the demonstration count and training budget load-bearing; the audit
promotes a wider 256-unit network as more reproducible than the single-seed best's 128.}
\label{fig:hero_rpo}
\end{figure}

On the docking problem the agent ran a single blind-mode campaign of \rpoIters{}
experiments. The story parallels the rendezvous, with the stakes raised: the policy must
now respect a hard keep-out constraint, so the audit certifies accuracy and safety at
once. The agent's advantage is no longer just that it docks more accurately; the
undirected baseline never produces a single feasible policy, while the agent is
strict-feasible on every reseed. Figure~\ref{fig:hero_rpo} summarizes the search.

\begin{figure}[tbp]
\centering
\includegraphics[width=0.86\textwidth]{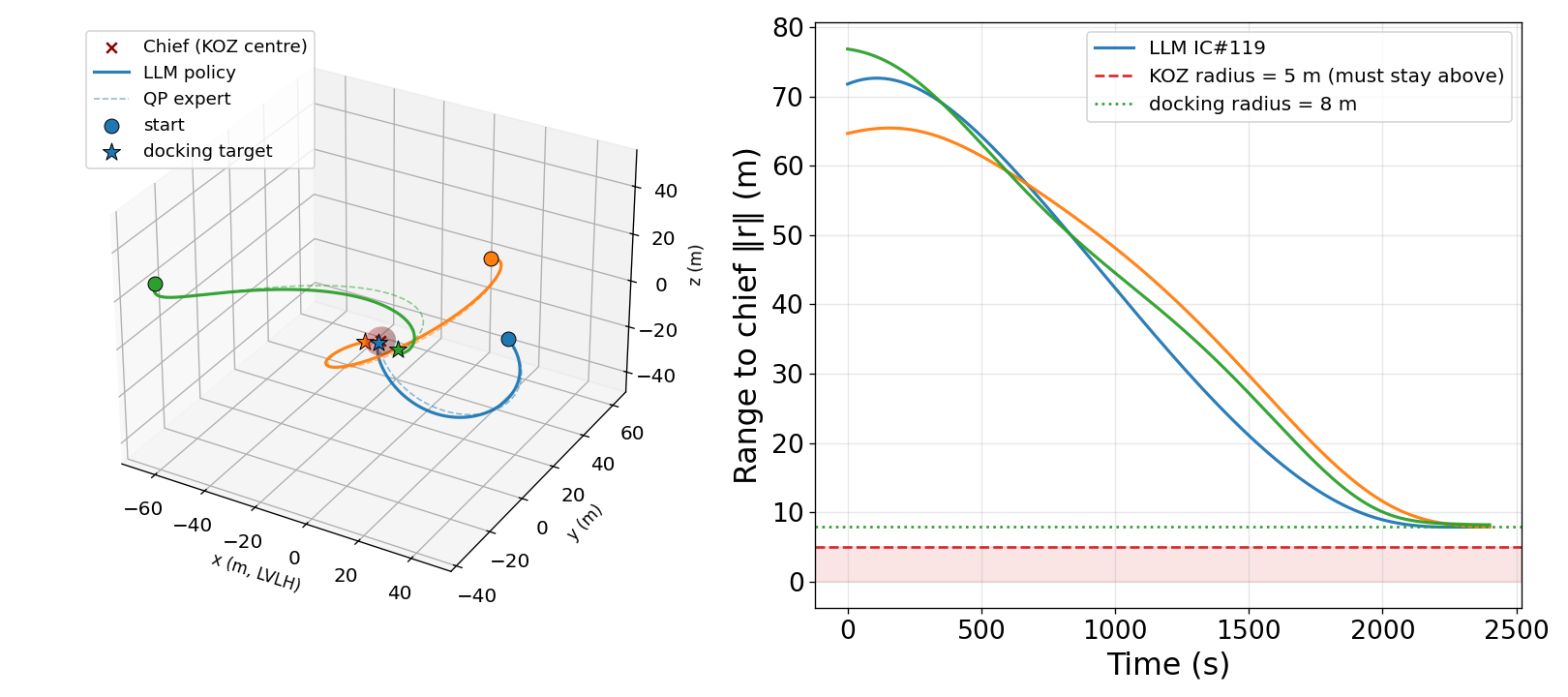}
\caption{The learned policy docks while respecting the keep-out zone. From held-out
far-side initial conditions, the cloned policy (solid) detours around the keep-out
sphere (shaded) and reaches the docking point, overlaying the optimal-control solution
(dashed); the minimum keep-out clearance stays nonnegative on every rollout.}
\label{fig:traj_rpo}
\end{figure}

From a default configuration that docks poorly but stays outside the keep-out zone, the
agent works the score down in stages to a single-seed best of \rpoSingle{}, well inside
the family target of \rpoTarget{}. As in the rendezvous, the search is methodical, not a random walk, with the run-time
CBF filter held on throughout, and the same intuition applies. Figure~\ref{fig:traj_rpo}
shows what the configuration buys in the state
space: starting on the far side of the chief, the cloned policy detours around the
keep-out sphere and docks, its trajectories overlaying the optimal-control expert's.

The same gate that certified the rendezvous now carries a second burden, to certify
safety alongside accuracy. The measured seed noise is \rpoFloor{}~$\pm$~\rpoFloorSd{} over
ten seeds. Reseeding the agent's best configuration over ten fresh seeds, disjoint from
those that measured it, yields \rpoLLM{}~$\pm$~\rpoLLMsd{}. Crucially,
\emph{every one of the ten reseeds is strict-feasible}: a success rate above the gate,
a mean violation rate of exactly zero, and a nonnegative minimum keep-out clearance.
As in the first family, the effect is reported directly: the reseeded mean improves on
the calibrated comparator by a difference of \rpoDelta{} (95\% confidence interval
\rpoDeltaCI), which is $m\approx\rpoSigmaMargin$ units of the measured noise standard
deviation, far beyond the two-sigma acceptance threshold. Both halves of the gate, the effect-size margin and strict feasibility on every seed,
are met.

As in the rendezvous family, reseeding also revises which configuration to designate as best. The
campaign's best single run used a narrower network (hidden width 128, reaching
\rpoSingle{} at its seed). But under ten fresh seeds that network is erratic (mean
$0.61$, standard deviation $0.24$), whereas a wider network (hidden width 256) is far
more reproducible (\rpoLLM{}~$\pm$~\rpoLLMsd{}) and is promoted as the family best.
Table~\ref{tab:rpo} collects the arms.

\begin{table}[t]
\centering
\caption{Credibility-audited results on the collision-avoidance docking problem.
Lower score (Eq.~\eqref{eq:rpo_score}, meters-equivalent) is better; the
optimal-control expert reaches the dock essentially exactly. Strict feasibility requires success
$\ge 0.80$, zero KOZ violations, and nonnegative clearance on every seed.}
\label{tab:rpo}
\small
\begin{tabular}{l c c c}
\toprule
Arm & Score & Strict-feasible & vs.\ seed noise \\
\midrule
Seed noise (n=10) & $\rpoFloor \pm \rpoFloorSd$ & no (no dock) & reference \\
Random search (best of \rpoRandN{}) & $\rpoRand$ & \textbf{0 of \rpoRandN{}} & not gated \\
LLM agent, single-seed best     & $\rpoSingle$ & yes & not gated \\
LLM agent, reseeded (n=10)      & $\mathbf{\rpoLLM \pm \rpoLLMsd}$ & \textbf{10 of 10} & $\mathbf{\rpoSigmaMargin\sigma}$, \textbf{PASS} \\
\bottomrule
\end{tabular}
\end{table}

On this problem the gap between directed and undirected search is categorical, not
merely quantitative. Over \rpoRandN{} trials sampling the same editable parameters,
random search never produces a single strict-feasible policy. Its best score is
\rpoRand{}, more than an order of magnitude above the agent's reseeded mean, and even
that is the single luckiest draw rather than the start of a trend. The rest of the
cloud sits at tens to hundreds of meters, closed-loop policies that drift away from
the target rather than docking. Two distinct failures keep every random configuration
out of the gate. First, random search leaves the run-time safety filter switched off
in more than half of its draws. Every draw that turned it off penetrated the
keep-out zone, while every draw that happened to leave it on stayed clear; this is
why the on/off switch is the load-bearing safety lever. Second,
even the filter-on draws never learn to dock, so their success rate never approaches
the gate's threshold. Either way, not one random configuration is strict-feasible. The
agent, by contrast, is strict-feasible on all ten reseeds: it
learns both to keep the run-time filter engaged and to imitate the detouring expert.
Panel~(a) of Fig.~\ref{fig:hero_rpo} shows the two best-so-far traces over the cloud
of raw per-experiment scores.

Finally, leave-one-out pruning reverts each edit in turn and remeasures against the measured
seed noise. Two edits prove load-bearing: the number of demonstrations and the
training budget. Reverting the demonstration count alone returns the score to about
$6.6$, back near the baseline, while batch size, weight decay, learning rate,
hidden width, and depth carry no individual signal. A distilled recipe that keeps
only those two edits and reverts the rest, including the network back to its default
width, still clears the seed noise by about five standard deviations, still meets
the family target, and remains strict-feasible on every verification seed. The win
therefore distills to a simple recipe, more demonstrations and more training,
rather than an elaborate architecture. The wider network and tuned learning rate the
agent also adopted improve the reseeded mean and its reproducibility, which is why the
audit designates the wide network at \rpoLLM{} as the family best, though neither is
required to clear either bar. At a coarser,
small-sample seed noise the learning rate had looked load-bearing; it dropped out
once the noise was remeasured over ten seeds, a reminder that pruning resolves only
effects that are large relative to the measured noise.

%% file: sections/07_conclusion.tex
\section{Conclusion and Future Work}
\label{sec:conclusion}

This paper presented AutoResearch, an agentic framework in which a large language model
autonomously drives the machine-learning research loop for aerospace control
problems, paired with a credibility layer that separates genuine progress from
seed luck. The framework's reusable family contract, a plain-language description,
one editable training script, a single structured metric, and an append-only run
log, lets the identical agent loop apply across very different problems, while
measured seed noise, reseeded verification, and leave-one-out pruning make each
headline result auditable. On the
Clohessy--Wiltshire rendezvous problem the agent reduced the terminal error of a
behaviorally cloned policy until it cleared the measured seed noise by a wide
margin, verified by reseeding over ten fresh seeds, while remaining
strict-feasible, and pruning then identified which of its edits actually carried
the result. On a second,
safety-constrained family, collision-avoidance docking past a keep-out zone, the
same machinery certified a strict-feasible policy that an undirected search could
not match, showing that the loop and its audit carry from one problem to another
despite the change in physics.

Several limitations bound these results. Each campaign runs for tens of iterations
rather than the thousands typical of large-scale automated search, trading throughput
for credibility per experiment, and each uses a single language-model backend whose
influence on the search is not yet characterized. The random-search comparison is a
secondary check rather than the headline: its sampling covered a smaller training
budget than the agent's winning configuration reached, so the central claim rests on
the seed-noise gate rather than on that comparison. Each result also comes from a
single agent run, so while reseeding controls for training-seed luck in the final
configuration, the run-to-run variance of the agent's own search is not yet
characterized.

Several directions remain. Future work will characterize sensitivity to the choice of
language-model backend and tighten the random-search arm to an exactly matched
sampling range. The broader aim is unchanged: to make LLM-driven experimentation a
dependable tool for developing and validating the learned autonomy that future
space missions will require, fast, and honest about what the evidence supports.